\documentclass[sn-mathphys,Numbered]{sn-jnl}

\usepackage{graphicx}%
\usepackage{multirow}%
\usepackage{amsmath,amssymb,amsfonts}%
\usepackage{amsthm}%
\usepackage{mathrsfs}%
\usepackage[title]{appendix}%
\usepackage{xcolor}%
\usepackage{textcomp}%
\usepackage{manyfoot}%
\usepackage{booktabs}%
\usepackage{algorithm}%
\usepackage{algorithmicx}%
\usepackage{algpseudocode}%
\usepackage{listings}%

\usepackage[T1]{fontenc}

\begin{document}

\title[Article Title]{Basketball-SORT: An Association Method for Complex Multi-object Occlusion Problems in Basketball Multi-object Tracking}  

\author[1]{\fnm{Qingrui} \sur{Hu}}\email{hu.qingrui@g.sp.m.is.nagoya-u.ac.jp}
\author[1]{\fnm{Atom} \sur{Scott}}\email{atom.james.scott@gmail.com}
\author[1]{\fnm{Calvin} \sur{Yeung}}\email{yeung.chikwong@g.sp.m.is.nagoya-u.ac.jp}
\author*[1,2,3]{\fnm{Keisuke} \sur{Fujii}}\email{fujii@i.nagoya-u.ac.jp}

\affil[1]{\orgdiv{Graduate School of Informatics}, \orgname{Nagoya University}, \orgaddress{\street{Chikusa-ku}, \city{Nagoya}, \state{Aichi}, \country{Japan}}}

\affil[2]{\orgdiv{RIKEN Center for Advanced Intelligence Project}, \orgname{1-5}, \orgaddress{\street{Yamadaoka}, \city{Suita}, \state{Osaka},  \country{Japan}}}

\affil[3]{\orgdiv{PRESTO}, \orgname{Japan Science and Technology Agency}, \orgaddress{\city{Kawaguchi}, \state{Saitama},\country{Japan}}}

\abstract{Recent deep learning-based object detection approaches have led to significant progress in multi-object tracking (MOT) algorithms. The current MOT methods mainly focus on pedestrian or vehicle scenes, but basketball sports scenes are usually accompanied by three or more object occlusion problems with similar appearances and high-intensity complex motions, which we call complex multi-object occlusion (CMOO). Here, we propose an online and robust MOT approach, named Basketball-SORT, which focuses on the CMOO problems in basketball videos. To overcome the CMOO problem, instead of using the intersection-over-union-based (IoU-based) approach, we use the trajectories of neighboring frames based on the projected positions of the players. Our method designs the basketball game restriction (BGR) and reacquiring Long-Lost IDs (RLLI) based on the characteristics of basketball scenes, and we also solve the occlusion problem based on the player trajectories and appearance features. Experimental results show that our method achieves a Higher Order Tracking Accuracy (HOTA) score of 63.48$\%$ on the basketball fixed video dataset and outperforms other recent popular approaches. Overall, our approach solved the CMOO problem more effectively than recent MOT algorithms. 
}


\maketitle

\section{Introduction}\label{Introduction}
    Multi-object tracking (MOT) is a fundamental computer vision task that aims to track multiple objects in a video and localize them in each frame. It involves the simultaneous detection, localization, and tracking of multiple objects in a video sequence. It has become increasingly important in various fields such as autonomous driving \cite{geiger2012we}, surveillance \cite{dendorfer2020mot20, milan2016mot16}, and sports analysis \cite{wang2022sportstrack}, most recent tracking algorithms which mainly focus on crowded street scenes \cite{dendorfer2020mot20, milan2016mot16}, static dancing \cite{sun2022dancetrack}, driving scenarios \cite{geiger2012we} and sports analysis \cite{wang2022sportstrack}. 
    
    Currently, the most popular approach for MOT is tracking-by-detection. The process of the tracking-by-detection MOT method \cite{bewley2016simple, peng2020tpm, aharon2022bot} usually consists of the following steps: (1) Object detection: First, an object detection algorithm is used to detect the target in each frame. (2) Object association: The targets detected in consecutive frames are associated. The association is achieved by matching the target detected in the current frame with the previously tracked target, such as using the Kalman filter (KF), which is famous for this task. (3) Object tracking: The tracking process begins once the targets have been associated between frames. This involves estimating the trajectory and state of each target over time. These current mainstream methods \cite{geiger2012we,liang2022rethinking, zhang2022bytetrack} have achieved good results on several open-source datasets, which are primarily in pedestrian and driving scenarios. However, these methods do not perform well on challenging datasets, especially on some datasets with designed sports scenarios \cite{cui2023sportsmot} since The tracking problem for sports scenarios is more complex. Nowadays, the demand for sports performance analysis in sports is growing, so the field of sports multi-target tracking needs more attention.

\begin{figure}[htbp]
    \centering
    \includegraphics[width=1.0\textwidth]{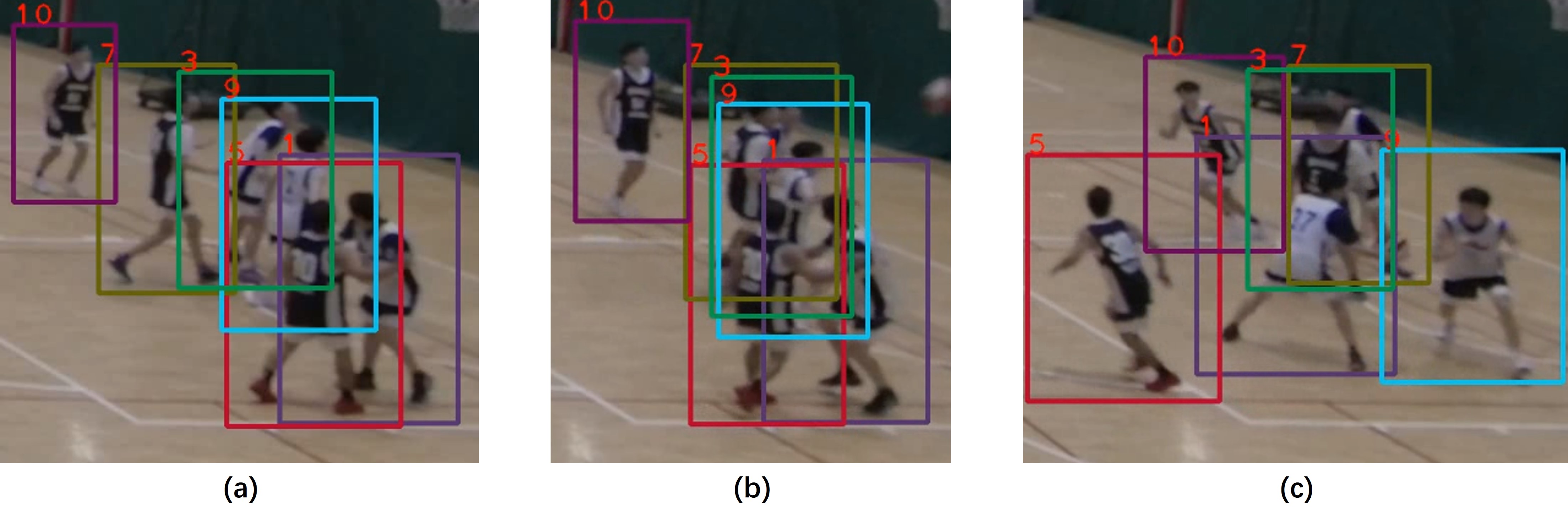}
    \caption{An example of Complex Multi-object Occlusion (CMOO) problems with a sequence of (a) to (c), including three or more player occlusions with similar appearance of players' jerseys in the same team and high-intensity complex and unpredictable motion. This can greatly affect detection and tracking accuracy and will cause a player to be lost and an ID switch.}
    \label{fig: CMOO}
    \end{figure}

    Unlike MOT for pedestrians or vehicles, team sports face more challenges due to a variety of reasons, including three or more object occlusions with similar appearance and high-intensity complex and unpredictable motions as illustrated in Figure \ref{fig: CMOO}, which we call Complex Multi-object Occlusion (CMOO) problems that frequently appear in basketball. Previous methods used appearance-motion fusion \cite{zhang2021fairmot, wojke2017simple} or simply motion-based methods \cite{zhang2022bytetrack, cao2023observation} to solve the association problems, but they did not focus on CMOO problems.

    In this paper, we propose a robust online MOT algorithm specifically designed for solving the CMOO problems, called Basketball-SORT, which is inspired by the SORT (Simple Online and Realtime Tracking) algorithm \cite{wojke2017simple}. Among various team sport MOT datasets \cite{cioppa2022soccernet, scott2022soccertrack, cui2023sportsmot, scott2024teamtrack}, we use the basketball fixed camera dataset from the TeamTrack dataset \cite{scott2024teamtrack} to validate our approach because it includes frequent CMOO problems. Our experimental results show that our algorithm effectively solves the CMOO problems and outperforms all other tracking algorithms on the basketball fixed camera dataset.
    
    Our paper has three main contributions as follows. 
    (1) We propose an online and robust MOT approach, named Basketball-SORT, which solves the CMOO problems, including three or more object occlusions with similar appearance of objects and high-intensity and unpredictable motions. 
    (2) We have addressed the frequent occlusion problems in basketball scenes by utilizing the Re-Identification (ReID) features, positions, and velocities of players before and after occlusions. By incorporating BGR and RLLI, we can resolve the issue of players losing their IDs after being occluded for an extended period.
    (3) Experimental results show that our method achieves a Higher Order Tracking Accuracy (HOTA) score of 63.48$\%$ on the basketball fixed video dataset and outperforms other recent popular approaches. Overall, our approach solved the CMOO problem more effectively than recent MOT algorithms.

\section{Related work}\label{Related work}
\subsection{Motion-based Multi-Object Tracking}
    Despite the significant advancements in object detection algorithms, many contemporary end-to-end MOT models still fall short in performance compared to traditional motion model-based tracking techniques. The Kalman filter \cite{kalman1960contributions} serves as the cornerstone for the most famous family of tracking-by-detection approaches.
    SORT \cite{bewley2016simple} used linear motion mode with IoU (intersection-over-union)-associated motion trajectory. ByteTrack \cite{zhang2022bytetrack} used the low score detection frame to predict the missing pedestrians, achieving good performance by balancing the detection quality and tracking confidence. Recently, OC-SORT \cite{cao2023observation} improves correlation accuracy for nonlinear motion scenes using an observation-centered approach. Some methods utilize the bounding box distance as the cost of the association, while some recent works utilize different IoU computation methods, such as calculating the BIou (the buffer of two overlapping boxes) \cite{yang2023hard}. Another work iteratively expanding the IoU according to different scales of expansion (EIoU) \cite{huang2024iterative} for the bounding box association between frames, which also demonstrates the effectiveness of MOT in SportsMOT \cite{cui2023sportsmot} and SoccerNet-Tracking \cite{cioppa2022soccernet} dataset. 
     
\subsection{Appearance-based Multi-Object Tracking}
    Visual identification serves as an intuitive cue for associating targets over time. One of the pioneering methods to incorporate deep visual features for object association is DeepSORT \cite{wojke2017simple}. Several approaches \cite{wojke2017simple, aharon2022bot} employ ReID models to extract embedding features from detections, which are then used for association. In recent years, the emergence of transformers \cite{vaswani2017attention} has sparked a new trend in utilizing appearance for MOT, where object association is formulated as a query-matching task \cite{cao2022track, zeng2022motr, sun2020transtrack}. However, it has been observed that appearance-based methods tend to be less effective when the objects of interest have similar visual characteristics \cite{sun2022dancetrack} or are subject to occlusion \cite{dendorfer2020mot20}, which is often the case in basketball scenarios.

\subsection{Multi-Object Tracking in Sports}
    Multiple Object Tracking (MOT) in sports environments presents significantly greater challenges compared to other domains where MOT is applied. This can be attributed to the unique characteristics of sports, such as the rapid and unpredictable movements of athletes, the visual similarity among players within the same team, and the increased occurrence of occlusions due to the dynamic nature of the sport.
    Several researchers have made notable contributions to address these challenges in various sports. For instance, in hockey, Vats et al. \cite{vats2023player} propose a method that integrates team classification and player identification techniques to enhance tracking performance. In football, Maglo et al. \cite{maglo2022efficient} showcase improved tracking accuracy by localizing the field and players. Moreover, Sang\"{u}esa et al. \cite{sanguesa2019single} leverage human pose information and actions as embedding features to boost the tracking performance of basketball players. Huang et al. \cite{huang2023observation} tackle multiple sports scenarios, including basketball, volleyball, and football, by combining OC-SORT with appearance-based post-processing.
    Our method primarily focuses on the CMOO problem in basketball. By incorporating CMOO's characteristics, we aim to perform robust and accurate player tracking.

\section{Methods}\label{Methods}
\begin{figure}[htbp]
    \centering
    \includegraphics[width=1.0\textwidth]{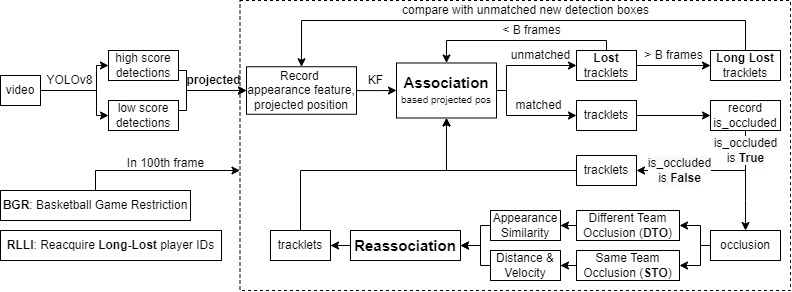}
    \caption{The pipeline of our Basketball-SORT algorithm. The main technical contributions are BGR and RLLI. BGR: In a basketball game, there can only be 10-player motion trajectories. At the 100th frame, we calculate the 10 longest trajectories on the court and identify them as players. RLLI: For the Long-Lost state player, we compare the position and appearance information of the Long-Lost player with the new detection bounding box to determine whether the player tracking ID should be reacquired.}
    \label{fig: pipeline}
    \end{figure}

    Our approach adopts the tracking-by-detection paradigm, which also enables online tracking without using future information. Our method has the following three steps, as shown in Figure \ref{fig: pipeline}. In Section \ref{subsec: projection}, We first use a fine-tuning yolov8 model to detect all the players. We then project the players' image position information onto the 2D court plane and associate all trajectories of each frame based on the players' 2D position information. Second, in Section \ref{subsec: BGR and RLLI}, we set the basketball game restriction (BGR) according to the rules of the basketball game and reacquire the tracking ID of the player in the Long-Lost state (RLLI) according to the trajectory characteristics of the player. Third, in Section \ref{subsec: STO and DTO}, due to BGR and the RLLI, the ID-increasing problem is converted to the ID-switch problem. A player's occlusion causes the ID-switch, and we use appearance features and motion features of trajectories to solve the ID-switch problem.
    
\subsection{Trajectory tracking based on projected position}\label{subsec: projection}
\begin{figure}[htbp]
    \centering
    \includegraphics[width=0.8\textwidth]{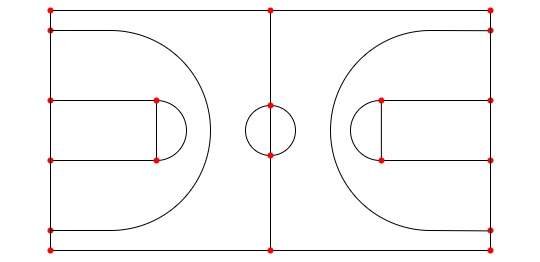}
    \caption{Standard basketball court projection, We use the red dots in the figure above to convert the image coordinate system coordinates to the player's position in the basketball court coordinate system}
    \label{fig: keypoint}
    \end{figure}

     Among various team sport MOT problems \cite{cioppa2022soccernet, scott2022soccertrack, cui2023sportsmot, scott2024teamtrack}, we consider the basketball fixed camera problems from the TeamTrack dataset \cite{scott2024teamtrack} to validate our approach because it includes frequent CMOO problems. Since the camera is fixed, we find the image coordinates of these basketball court key points in the video based on the 20 key points in the standard overhead view of the court as illustrated by the red dots in Figure \ref{fig: keypoint}, and then calculate the homography matrix by the following formula:
    \begin{equation}
        \left[\begin{array}{l}
            x_{c} \\
            y_{c} \\
            w_{c}
        \end{array}\right] = H
        \left[\begin{array}{l}
            x_{i} \\
            y_{i} \\
            1
        \end{array}\right]
        \label{eq: homography}
    \end{equation}
    where $(x_{c}, y_{c})$ denotes the coordinates of the basketball court coordinate system, and $(x_{i}, y_{i})$ denotes the coordinates of the image coordinate system, $w_{c}$ denotes the scale factor which can unifying projection transformations into matrixes. The homography matrix $H$ can be computed from the coordinates of the key points of the 20 standardized courts we manually input (Figure \ref{fig: keypoint}).

    We use the bottom midpoint of the detection bounding box as the player's image coordinate system position to convert it into a top-view projection position.
    As a result, we can get the relative coordinates of the players in the court's plane, which could be a better help in tracking the player's position in the court. The player is associated with each frame using the Kalman filter based on the position and speed of the projected position moving on the plane. We did not utilize the IoU association because of the frequent occurrence of CMOO where players are far apart in actual distance but have completely overlapping IoUs. The reason for not employing ReID for association is due to the appearance similarity caused by the players' uniforms.

\subsection{Basketball Game Restriction and Reacquiring Long-Lost IDs}\label{subsec: BGR and RLLI}
    In a basketball game, there is often severe occlusion of players due to offense and defense, which often leads to the lost player being assigned a new tracking ID after reappearing. 
    In addition, due to the detector's limited performance and the confusion between referees and players, some referees and the audience may be detected as players and have some redundant trajectories. 
    To solve these problems, we added the basketball game restriction (BGR): because basketball games have 10 players, there will only be 10 tracking IDs in a game, solving the problem of the IDs increasing.
    We calculate the length of all trajectories at the designed frame and select the top ten trajectories with the most extended lengths in the court to be recognized as players. In this study, this designated frame is set at the 100th frame because referee and audience trajectories are shorter, and all 10 player trajectories are available at the 100th frame. These ten trajectories will be fixed, and no new trajectories will be generated.
    
    When 10 players' trajectories are determined, if a trajectory is occluded for less than $B$ frames, we define this trajectory as a Lost state. If a trajectory is lost for more than $B$ frames, we define this trajectory as a Long-Lost state.
    Because players usually play 1-on-1, the resulting trajectory states are often defined as Lost or Long-Lost states.
    For the Lost state player, we still use the KF to predict each frame's real-time position and detection box matching.
    For the Long-Lost state players, we need to reacquire their tracking ID (RLLI). We record the player's court position, and ReID features before the player loses. When a new detection reappears on the court, we reassign the player's ID by comparing the appearance similarity and distance before disappearing. 
    The appearance similarity is a vital clue for object association between frames, which can filter out some impossible associations. It can be calculated by the cosine similarity between the appearance features, and we will compute the appearance feature for each frame of the detection box. The cost for appearance association ${ Cost }$ can be directly obtained from the cosine similarity with the following formula:
    \begin{equation}
       \text { Cost }_{a\_b}=1-\text { Cosine Similarity }=1-\frac{a \cdot b}{\|a\|\|b\|}
       \label{eq: cosine similarity}
    \end{equation}    
    where $a$ and $b$ denote the appearance characteristics of the detection box for two specific frames, respectively. A higher cosine similarity indicates a higher appearance similarity, while a lower cosine similarity indicates that the appearance of the trajectory is different from the detection appearance. 
    The reappeared detection of whether to rematch the lost trajectory needs to meet the following conditions:
    \begin{equation}
        \begin{cases}
            { Cost }_{lost\_re} > \alpha \\
            { Dist }_{lost\_re} > \beta  
        \end{cases}
        \label{eq: long lost state}
    \end{equation}
    where ${ Cost }_{lost\_re}$ denotes the cosine similarity between the lost trajectory's appearance feature before disappearing and the reappeared detection appearance feature; $Dist_{lost\_re}$ represents their Euclid distance; $\alpha$ and $\beta$ denote the thresholds set for similarity and distance respectively.
    
\subsection{Solving Same and Different Team Player Occlusion Problem}\label{subsec: STO and DTO}
    In the previous SOTA tracking method \cite{huang2024iterative, wang2022sportstrack} in SportsMOT \cite{cui2023sportsmot} dataset, when a trajectory is lost for more than a specific frame, it discards the trajectory, and when a new detection frame with no trajectory matches reappears, it initializes the detection box as a new trajectory, which is called the ID-increasing problem.
    Since we added BGR, the ID-increasing problem has become the ID switch problem. In addition, the long-time occlusion caused by the defense, as mentioned in the previous section, will also lead to the ID switch.

    For the ID switch problem, first, we need to detect whether the player is occluded or not in each frame to record the occlusion state of the player. If the player's trajectory is lost, it means the player has been occluded. When the player trajectory reappears, we record the two players closest to the player in the Lost state in the occluded frame, indicating that the occluded player may have had an ID switch with these two players. In addition, we recorded the projected court position history and the ReID features for each frame.
    Then, we need to categorize the occlusion into Same Team Occlusion (STO) and Different Team Occlusion (DTO) according to whether the two occluded players are on the same team or not. We can determine which situation the occlusion belongs to based on the ReID similarity of the two players before and after the occlusion. The formula is as follows:
    \begin{equation}
        Occlusion =
        \begin{cases}
        STO,  \: if \: \left | R_{a\_N}^{before} - R_{b\_N}^{before} \right | < \gamma \: or \: \left | R_{a\_M}^{after} - R_{b\_M}^{after}  \right | < \gamma  \\
        DTO,  \: if \: \left | R_{a\_N}^{before} - R_{b\_N}^{before} \right | > \gamma \: or \: \left | R_{a\_M}^{after} - R_{b\_M}^{after}  \right | > \gamma 
        \end{cases}
        \label{eq: STO or DTO}
    \end{equation}
    where $R_{a\_N}^{before}$ denotes the ReID features of player $a$ for $N$ frames before occlusion occurs, and $R_{a\_M}^{after}$ denotes its ReID features for $M$ frames after occlusion occurs. The same can be applied to player $b$,  $\gamma$ denotes the similarity threshold between two players before and after occlusion, which is used to distinguish whether the occlusion is STO or DTO.

    Since we record the player's position, the ReID features, and the occlusion state information, we compare whether the IDs of the two players after occlusion are correctly assigned after the $M$ frame.
    For DTO, the different team uniforms will have lower appearance similarity, so we compare the ReID feature similarity of two trajectories before and after occlusion. If the ReID feature of one trajectory after occlusion is similar to the other one before occlusion, it is considered that an ID switch has occurred, and the formula is as follows:
    \begin{equation}
        \begin{cases}
            \left | R_{a\_N}^{before} - R_{b\_M}^{after}  \right | < \delta \\
            \left | R_{a\_N}^{after} - R_{b\_M}^{before} \right | < \delta 
        \end{cases}
        \label{eq: DTO}
    \end{equation}
    where $R_{a\_N}^{before}$ denotes the ReID feature of $N$ frames before the occlusion of player $a$, $R_{b\_M}^{after}$ denotes the ReID feature of $M$ frames after the occlusion of player $b$, $\delta$ denotes the threshold of appearance similarity of two players.  Because two players typically do not change much in their appearance features before and after occlusion occurs, Satisfying the Eq. \eqref{eq: DTO} means that these two players have wrongly assigned their IDs after the occlusion and need to exchange their IDs.
    
    For STO, the same team uniforms will have a higher degree of appearance similarity, so we compare the moving distance and speed of two trajectories before and after occlusion. 
    If the moving distance and velocity of one trajectory after occlusion are similar to the other one before occlusion, it is considered that an ID switch has occurred. The formulae are as follows:
    \begin{equation}
        \begin{cases}
        \left | V_{a\_N}^{before} - V_{a\_M}^{after} \right | > \varepsilon \\
        \left | V_{b\_N}^{before} - V_{b\_M}^{after} \right | > \varepsilon \\
        \left |(P_C^{occ} - P_{a\_N}^{before}) \right | - \left |(P_{a\_M}^{after} - P_C^{occ}) \right | > \zeta \\
        \left |(P_C^{occ} - P_{b\_N}^{before}) \right | - \left |(P_{b\_M}^{after} - P_C^{occ}) \right | > \zeta \\
        \end{cases}
        \label{eq: STO}
    \end{equation}
    where $V_{a\_N}^{before} = (\left | P_{C}^{occ} - P_{a\_N}^{before} \right |) / (\left | F_{C} - F_{N} \right |)$ and $V_{a\_M}^{after} = (\left | P_{a\_M}^{after} - P_{C}^{occ} \right |) / (\left | F_{M} - F_{C} \right |)$. $P_{C}^{occ}$ denotes that only one player's position can be detected during occlusion, $F_{C} - F_{N}$ denotes the number of frames from $N$ to $C$, $F_{M} - F_{C}$ denotes the number of frames from $C$ to $M$. $V_{a\_N}^{before}$, $V_{a\_M}^{after}$ denotes the velocity of player $a$ and $b$ at $N$ frames before occlusion and $M$ frames after occlusion, respectively. $V$ denotes the velocity and the unit is cm/frame, and $P$ denotes the position. $\varepsilon$ and $\zeta$ represent the thresholds for the velocity and position of two players, respectively. Since both players don't typically make large changes in velocity and position at the same time when the occlusion occurs, satisfying the Eq. \eqref{eq: STO} means that these two players could have wrongly assigned their IDs after the occlusion and need to exchange their IDs. 

\section{Experiments}\label{Experiments}
\subsection{Experimental setup}
    \noindent \textbf{Datasets.} We used the basketball fixed camera videos from the TeamTrack dataset \cite{scott2024teamtrack} to validate our approach because it is considered to include frequent CMOO problems among various team sport MOT datasets \cite{cioppa2022soccernet, scott2022soccertrack, cui2023sportsmot, scott2024teamtrack}. Here, we briefly introduce the basketball fixed video dataset, which was filmed from the side view using a fish-eye camera. The published videos were calibrated in advance and segmented into 30-second intervals. We split it into training and test videos with 162,353 and 170,529 frames, respectively. In the dataset, the players and referees were labeled, which allows our detector to detect only the players and filter out the referees.

    \noindent \textbf{Metrics.} Multiple object tracking accuracy (MOTA) \cite{bernardin2008evaluating} have been often used as an evaluation metric for MOT tasks. However, MOTA focuses more on detection performance rather than association accuracy. In basketball, we track mainly to get the player's trajectory, so we need to use HOTA \cite{luiten2021hota} as the evaluation metric. HOTA consists of detection accuracy (DetA), localization accuracy (LocA), and association accuracy (AssA), the metrics combined to evaluate detection accuracy and tracking accuracy. The metrics combine different aspects of the MOT and HOTA to reflect the system's performance more comprehensively. In addition, false positives (FP), false negatives (FN), and ID switches (IDS) can also reflect the tracking performance well. Therefore, we use HOTA, DetA, LocA, AssA, FP, FN, and IDS as the evaluation metrics of tracking.

    \noindent \textbf{Detector and ReID.} In our method and baselines, we chose YOLOv8 \cite{Jocher_YOLO_by_Ultralytics_2023} as our object detector to achieve real-time and high-accuracy detection performance. We used the officially provided yolov8l pre-training model to train our model with the training dataset. 
    We used an SGD optimizer with a weight decay of 0.0005 and momentum of 0.9. The initial learning rate is 0.001 and training for 200 epochs. For player ReID, we use the proposed FastReID model \cite{he2023fastreid} because we use ReID mainly to solve the occlusion problem, and the performance of the ReID model does not have high performance. All the experiments are conducted on a single Nvidia RTX 4080 GPU.

    \noindent \textbf{Implementation details. } The threshold for a detection to be treated as a high-score detection is 0.6. Detections with a confidence score between 0.6 and 0.1 will be treated as low-score detections, and the rest with a confidence score lower than 0.1 will be filtered. 
    We performed hyperparameter optimization on our method and found the best threshold. RLLI threshold $\alpha$ was set to 0.2 and $\beta$ set to 260; Distinguishing between STO and DTO parameters $\gamma$ was set to 0.2; DTO appearance similarity threshold $\delta$ was set to 0.2; DTO velocity similarity threshold $\varepsilon$ was set to 3 and position threshold $\zeta$ was set to 3.

\subsection{Benchmark Results}\label{sec: Benchmark results} 
\begin{figure}[htbp]
    \centering
    \includegraphics[width=1\textwidth]{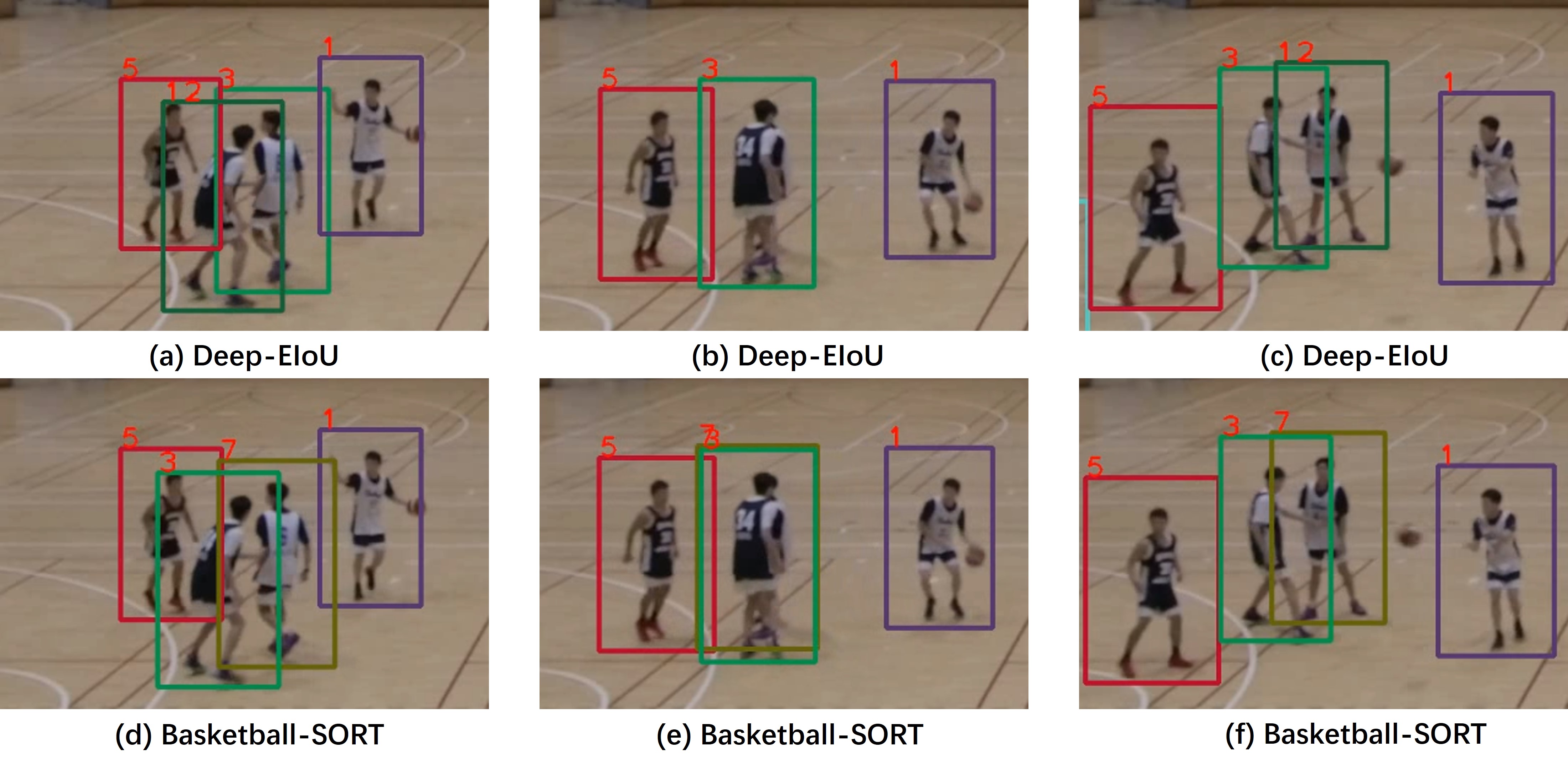}
    \caption{(a-c) show the results using the DeepEIoU method. During occlusion, the player ID is lost, and the method fails to assign the player ID correctly. (d-f) demonstrate the results using the Basketball-SORT approach, which successfully matches the player ID during complete occlusion.}
    \label{fig: compare}
    \end{figure}

    Among current MOT algorithms, BoT-SORT \cite{aharon2022bot} demonstrates the most representative performance on large-scale pedestrian datasets such as MOT20 \cite{dendorfer2020mot20}. On the other hand, Deep-EIoU \cite{huang2024iterative} achieves the best results on sports sense datasets like SportsMOT \cite{cui2023sportsmot}. Therefore, we compare our method with recent methods such as BoT-SORT and Deep-EIoU on the basketball fixed video dataset, We have also included an example, as depicted in Figure \ref{fig: compare}.
    The results are shown in Table \ref{tab:tabcompare}. 
    In these methods, we employed the same detector, YOLOv8, for all experiments. The results show that DetA and LocA were quite similar, fluctuating within 1 percentage point. Therefore, we will omit the DetA and LocA results in the subsequent sections.
    The results demonstrate that our method achieves the best HOTA score due to a significant improvement in AssA. By utilizing projected positions to associate player trajectories, our approach can correctly track players even when their projected positions are far apart but their detection bounding boxes overlap. Since this situation occurs frequently, it leads to a substantial reduction in IDS, resulting in a notable boost in AssA.
    When we employ BGR, we do not discard or create new trajectories. Not discarding trajectories prevents some lost player trajectories from requiring their correct IDs, resulting in an increase in False Negatives (FN). We also show it in the ablation study below.
    On the other hand, not creating new trajectories means that some detection bounding boxes of referees and spectators will not be recognized as player trajectories, thereby reducing False Positives (FP).
    In addition, our method yields only 10 player trajectories, which is significantly better than other methods.
    Our method achieves the best results and outperforms all the other previous trackers while keeping the tracking process online, showing our algorithm's effectiveness in MOT for the basketball fixed video dataset.

    \begin{table}[htbp]
    \scalebox{0.9}{
    \begin{tabular}{cccccc}
    \hline
    Method                     & HOTA (\%)           & AssA (\%)            & FN                   & FP                   & IDS       \\ \hline
    BoT-SORT                   & 59.70±6.43          & 53.43±10.05          & 270.8±163.3          & 492.6±369.9          & 23.2±13.7 \\ 
    Deep-EIoU                  & 60.74±7.72          & 55.47±11.62          & \textbf{265.0±162.4} & 486.2±368.1          & 20.7±10.3 \\ 
    Basketball-SORT(ours)      & \textbf{63.35±7.21} & \textbf{59.15±10.61} & 387.7±464.9          & \textbf{118.9±191.8} & \textbf{9.3±5.0}   \\ \hline
    \end{tabular}}
    \caption{The performance comparison among different MOT algorithms on the basketball fixed video test sets. Our algorithm outperforms all the other previous tracking algorithms and achieves the best performance result in several major evaluation metrics. BoT-SORT and DeepEIoU are evaluated based on their official code \cite{aharon2022bot, huang2024iterative}.}
    \label{tab:tabcompare}
    \end{table}

\subsection{Ablation Studies} 
    In this experiment, we evaluated Basketball-SORT (as the full model) on the test set of the basketball fixed camera dataset using different settings, including whether to use BGR, RLLI, and whether to employ STO and DTO to address occlusion issues (where ``mix'' indicates using both STO and DTO simultaneously). In addition, the ``Projected'' method, as mentioned in Section \ref{subsec: projection}, only utilizes the projection positions to associate the trajectories of players.
    The results are shown in Table \ref{tab: ablation}. We can see that after adding BGR, the FP and FN decrease dramatically, which may be due to the insufficient amount of data resulting in some audiences being recognized as players in the game. 
    The reasons for the increase in FN and decrease in FP caused by BGR are the same as in Section \ref{sec: Benchmark results}. STO and DTO lead to an increase in IDS because our approach to resolving occlusion issues involves identifying the occlusion after it occurs and reassigning the correct ID to the player, which also results in an additional IDS, thereby increasing the overall IDS count. 
    RLLI, STO, and DTO have all successfully improved the HOTA, indicating that occlusion problems in the sports scene occur frequently. 
    The results show that the STO+RLL method has the highest HOTA, probably because more of the occlusions in the dataset are STO, and some DTO are likely to be misclassified as STO.
    
    \begin{table}[htbp]
    \scalebox{0.75}{
        \begin{tabular}{cccccccccccc}
        \hline
        Method     & BGR         & RLLI       & STO        & DTO        & HOTA (\%)           & AssA (\%)            & FN                   & FP                   & IDS       \\ \hline
        Projected  &             &            &            &            & 61.96±6.99          & 57.32±10.07          & \textbf{239.6±142.7} & 532.2±409.7          & 20.1±10.3 \\
        BGR        & \checkmark  &            &            &            & 63.12±7.55          & 59.23±9.89           & 491.2±491.1          & 110.9±162.0          & \textbf{8.6±5.2}   \\
        RLLI       & \checkmark  & \checkmark &            &            & 62.35±7.02          & 58.50±9.59           & 500.2±425.9          & 156.5±194.4          & 8.9±5.1   \\
        STO        & \checkmark  &            & \checkmark &            & 63.45±7.74          & 59.57±10.67          & 442.9±527.8          & 105.6±151.4          & 8.9±5.1   \\
        STO + RLLI & \checkmark  & \checkmark & \checkmark &            & \textbf{63.48±7.10} & 59.37±10.29          & 386.6±465            & 117.5±192.5          & 9.0±4.8   \\
        DTO        & \checkmark  &            &            & \checkmark & 63.06±7.34          & 59.11±9.97           & 492.9±528.0          & 106.9±161.0          & 9.6±5.4   \\
        DTO + RLLI & \checkmark  & \checkmark &            & \checkmark & 63.18±6.75          & 58.74±9.80           & 386.5±464.1          & 117.7±193.3          & 9.4±5.2   \\
        mix        & \checkmark  &            & \checkmark & \checkmark & 63.21±7.84          & \textbf{59.91±11.01} & 492.7±527.2          & \textbf{105.1±161.0} & 9.7±5.5   \\
        Full model & \checkmark  & \checkmark & \checkmark & \checkmark & 63.35±7.21          & 59.15±10.61          & 387.7±464.9          & 118.9±191.8          & 9.3±5.0   \\ \hline
        \end{tabular}
    }
    \caption{We evaluate the Basketball-SORT algorithm with different settings on the basketball video test set. Including using the Basketball Game Restriction (BGR), Reacquiring Long-Lost IDs (RLLI), same team occlusion (STO), and different team occlusion (DTO) for our method.}
    \label{tab: ablation}
    \end{table}
    
\subsection{Robustness different movement velocity}\label{subsec: robustness}
    In basketball, the movement speed of players varies dynamically and is unpredictable. 
    When a player is occluded, the KF still predicts its possible position in the next frame using the match threshold. However, due to the instability of the player's velocity, using a fixed association threshold may result in the athlete being unable to rematch with a newly appeared detection bounding box after the occlusion. 
    To address this issue, we introduce the RLLI reacquire threshold ${ Dist }_{lost\_re}$ as mentioned in section \ref{subsec: BGR and RLLI}, which calculates the distance between the position of the Long-Lost trajectory and the reappearance of the detection frame.
    It can effectively reduce the probability that the Long-Lost player's trajectory cannot be re-matched with the new detection frame.
    In order to prove the robustness of our method, we use the RLLI model in Table \ref{tab: ablation} to test the effect of different match thresholds and ${ Dist }_{lost\_re}$ on the results as shown in Table \ref{tab: reacquire}.
    Match threshold represents the change in position due to velocity variations between adjacent frames. It is usually below 200 (cm), but in fast break situations, it may reach up to 300 (cm). Therefore, we chose this range of variations to test our method.
    ${ Dist }_{lost\_re}$ represents the distance between a Long-Lost player's trajectory and the reappeared detection bounding box within $B$ frames. We have set this parameter to a range of 170-250 (cm), as it corresponds to the most likely distance a Long-Lost player might move within the given $B$ frames.
    Since we have defined the court dimensions as 2800 cm × 1400 cm, the units for the Match threshold and ${ Dist }_{lost\_re}$ parameters are in centimeters (cm).
    The results show that different parameters give similar results of average HOTA $61.81\%$. This proves our method’s effectiveness in the real world, where ground truth is often not available and the tracking parameter can not be tuned.

    \begin{table}[htbp]
    \begin{tabular}{cccccccc}
    \hline
                                   & \multicolumn{6}{c}{Match threshold in KF} \\ 
                          &     & 200   & 220   & 240   & 260            & 280   & 300   \\ \hline
    ${ Dist }_{lost\_re}$ & 150 & 61.62 & 61.54 & 61.69 & 61.88          & 61.48 & 61.31 \\
                          & 170 & 61.89 & 61.71 & 61.87 & 61.90          & 61.34 & 61.48 \\
                          & 190 & 61.88 & 61.71 & 61.94 & 61.99          & 61.33 & 61.51 \\
                          & 210 & 61.84 & 61.93 & 61.90 & 62.34          & 61.75 & 61.48 \\
                          & 230 & 61.90 & 61.96 & 61.93 & 62.34          & 61.76 & 61.48 \\
                          & 250 & 61.91 & 61.96 & 61.93 & \textbf{62.35} & 61.78 & 61.48 \\ \hline
    \end{tabular}
    \caption{We tested different match thresholds, and RLLI reacquires thresholds in the basketball fixed video test set. Each row parameter of the table represents different ${ Dist }_{lost\_re}$, and each column parameter represents a different match threshold in KF. The values in the cells represent the HOTA, and various parameter configurations yield similar HOTA scores, indicating the robustness of the tracking method across different settings.}
    \label{tab: reacquire}
    \end{table}

\section{Conclusion}\label{Conclusion}
    In this paper, we propose Basketball-SORT, an online and robust MOT approach that solves the CMOO problems in basketball videos. 
    To overcome the CMOO problem, we used the trajectories of neighboring frames based on the projected positions of the players. Our method designs the BGR and RLLI based on the characteristics of basketball scenes, and we also solved the occlusion problem based on the player trajectories and appearance features.
    Experimental results show that our approach can effectively solve the CMOO problem and is much better than previous tracking algorithms.
    For future work, tracking basketball games filmed with moving cameras and addressing the more complex player occlusion issues in basketball scenes are needed in order to fully track the motion trajectories of athletes in more general ways.

\section*{Acknowledgments}
This work was financially supported by JSPS Grant Number 20H04075 and 23H03282 and JST PRESTO Grant Number JPMJPR20CA.

\section*{Declarations}
\subsection*{Conflict of Interest}
The authors declare that they have no conflict of interest.
\subsection*{Data availability statements}
The evaluation dataset in this study is available at \url{https://github.com/AtomScott/TeamTrack}. 

\subsection*{Compliance with Ethical Standards}
The participants were fully informed about the study, and their consent was obtained in advance. All the experimental procedures were performed after obtaining prior approval from the ethical committee at Tokai University.

\bibliography{sn-bibliography}

\end{document}